\documentclass[10pt,twocolumn,letterpaper]{article}

\usepackage[pagenumbers]{cvpr} 

\usepackage{algorithm}
\usepackage{listings}
\lstdefinestyle{geometrypython}{
  language=Python,
  basicstyle=\fontencoding{T1}\fontsize{8.5}{10}\selectfont\ttfamily,
  keywordstyle=\color{blue!45!black}\bfseries,
  commentstyle=\rmfamily\color{black!65}\itshape,
  stringstyle=\color{teal!60!black},
  numbers=left,
  numberstyle=\scriptsize\color{black!40},
  numbersep=6pt,
  xleftmargin=14pt,
  columns=fullflexible,
  keepspaces=true,
  showstringspaces=false,
  upquote=true,
  aboveskip=5pt,
  belowskip=4pt
}

\definecolor{cvprblue}{rgb}{0.21,0.49,0.74}
\usepackage[pagebackref,breaklinks,colorlinks,allcolors=cvprblue]{hyperref}

\def\paperID{*****} 
\def\confName{CVPR}
\def\confYear{2026}

\title{DVTR-KD: Distilling Vision-Language Models via Intermediate Visual Token Reconstruction}

\author{
Haolan Wei \quad Jiahui Wang \quad Junli Zha \quad Xinkai Lu \quad Yunxiao Zou \quad
Chao Zhou \quad Xiang Song \\ [4pt]
SF Technology Co., Ltd\\
Shenzhen, China\\
{\tt\small weihaolan6@gmail.com, } 
{\tt\small \{wangjiahui, zhajunli\}@sf-express.com}
}

\begin{document}
\maketitle
\begin{abstract}
Multimodal large language models (MLLMs) have achieved strong performance on diverse vision-language tasks, but their computational demands limit deployment under constrained budgets, particularly on edge devices.
Many works focus on Knowledge Distillation (KD) and try to transfer knowledge from powerful teacher models to students. In addition to reducing model size, visual token compression provides a complementary route to efficiency, but introduces mismatched visual token counts that complicate direct token-wise knowledge transfer.
To mitigate this, we propose DVTR-KD, a knowledge distillation framework based on \textbf{D}ense \textbf{V}isual \textbf{T}oken \textbf{R}econstruction, which reconstructs the teacher's dense visual tokens from the student's compressed tokens.
We reconstruct visual states after the final LLM layer and at the input to an intermediate layer identified as important for visual-information readout. At the intermediate layer, reconstruction is supervised through the teacher's response predictions rather than by direct token-wise KL or MSE matching.
An auxiliary attention-readout loss complements this response-level supervision by aligning how the student and the original teacher integrate contextual information when predicting responses.
Evaluations on diverse benchmarks demonstrate that DVTR-KD improves the performance of token-compressed students while retaining the inference efficiency afforded by visual token compression.
\end{abstract}    
\section{Introduction}
\label{sec:intro}

Large language models (LLMs)~\cite{qwen3,xu2026deepseek,zeng2026glm,grattafiori2024llama3,bai2023qwen} have achieved strong performance in language understanding, generation, and reasoning. Building upon these advances, recent studies have sought to extend LLMs beyond the text modality by integrating visual perception capabilities, giving rise to Multimodal Large Language Models (MLLMs). By connecting vision encoders with powerful LLMs \cite{llava,kimi_vl,kimik3,li2023blip,bai2025qwen3}, MLLMs are able to perceive visual content and perform language-based reasoning over multimodal inputs. This paradigm has achieved impressive performance across a wide range of vision-language tasks, including visual question answering \cite{antol2015vqa,vqav2,scienceqa,gqa}, image understanding \cite{liu2024mmbench}, optical character recognition \cite{docvqa,textvqa,ocrbench,chartqa}, and visual reasoning \cite{mmmu}, establishing MLLMs as a promising foundation for general-purpose multimodal intelligence.

However, unlike earlier MLLMs that process images at fixed resolutions \cite{llava,llava15,li2023blip,zhu2023minigpt}, advanced models increasingly support native-resolution visual inputs to preserve fine-grained visual information \cite{kimik3,bai2025qwen3}. While such designs improve visual perception, especially for high-resolution images in perception tasks, they often produce substantially longer visual token sequences and consequently introduce considerable computational cost. This issue becomes particularly pronounced in the LLM decoder, where the computational cost grows quadratically with the sequence length. To improve the inference efficiency, some researchers focus on reducing the number of visual tokens while preserving the capabilities of MLLMs \cite{llava_uhd,sun2025llavauhdv3}. These methods reduce visual redundancy, enabling MLLMs to process high-resolution visual inputs with significantly fewer tokens.

Knowledge distillation (KD) provides an effective way for transferring knowledge from a knowledgeable teacher model to a student model, typically by encouraging the student to imitate the teacher's output distributions or representations. In the context of MLLMs, several studies have explored knowledge transfer between teacher and student MLLMs. LLaVA-KD \cite{cai2025llava}, for example, presents a comprehensive MLLM distillation framework that combines conventional response distillation with visual representation and relation distillation. However, relatively little attention has been paid to distillation between a teacher and a token-compressed student, where the teacher and student produce visual sequences of substantially different lengths. EM-KD \cite{em-kd} addresses this mismatch using the Hungarian matching algorithm to establish one-to-one correspondences between teacher and student visual tokens. Despite its effectiveness, such one-to-one matching inherently utilizes only a subset of the teacher's dense visual tokens. Given $N_T$ teacher tokens and $N_S<N_T$ student tokens, one-to-one matching uses at most $N_S$ teacher tokens, leaving at least a fraction $1-N_S/N_T$ without direct token-level supervision. Consequently, an increasingly large fraction of the teacher's visual representations remains unmatched as the compression ratio increases, potentially limiting the fine-grained visual knowledge transferred to highly compressed students.

To address this limitation, we propose \textbf{DVTR-KD}, which takes different direction: rather than selecting a subset of teacher tokens to match the compressed student sequence, we reconstruct dense visual representations from the student's compressed visual tokens and align them with the complete teacher visual sequence. By performing token-wise distillation on the reconstructed representations, DVTR-KD allows supervision from all teacher visual tokens, encouraging the compressed student representations to preserve sufficient information to recover the teacher's fine-grained visual knowledge.

\begin{figure}[t]
    \centering
    \includegraphics[width=\linewidth]{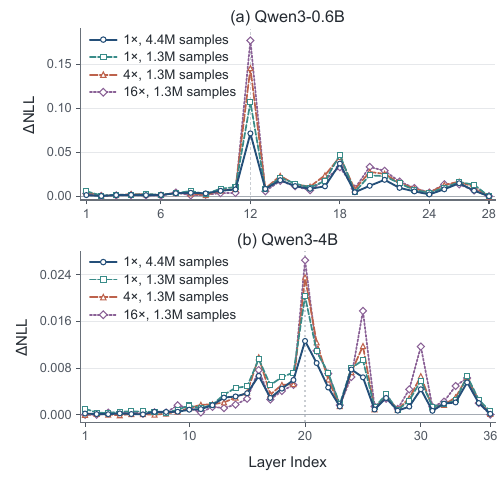}
    \caption{Layer-wise sensitivity to visual-value ablation after SFT and before distillation.
    We zero the value vectors of all visual tokens in one transformer layer at a time and report the change in response NLL ($\Delta\mathrm{NLL}$), averaged over the same diagnostic subset.
    In the legends, $1\times$ denotes uncompressed models, while $4\times$ and $16\times$ denote models with nominal fourfold and sixteenfold reductions in visual-token count.
    The sample counts are summed over all training datasets before distillation.
    }
    \label{fig:visual_readout_layers}
\end{figure}

In addition, while Align-KD \cite{align-kd} explores cross-modal distillation through first-layer attention and visual-feature alignment, the importance of visual information may vary across different depths of the LLM. To investigate this, we conduct a layer-wise visual ablation study on MLLMs built upon Qwen3 \cite{qwen3} and MoonViT \cite{kimi_vl}, trained under different compression ratios and quantities of training samples. Specifically, we zero out the value vectors of visual tokens in the self-attention module of one LLM layer at a time and measure the resulting increase in response $\mathrm{NLL}$. As illustrated in Figure~\ref{fig:visual_readout_layers}, removing visual information at the first layer has a relatively limited impact, whereas perturbations at certain intermediate layers lead to substantially larger degradation. We refer to the layer exhibiting the largest $\Delta\mathrm{NLL}$ as the \textbf{Principal Visual Readout Layer (PVRL)}. We observe that for a given LLM backbone, the PVRL remains same across different training data scales and visual token compression ratios. Meanwhile, stronger models consistently exhibit smaller performance degradation under the same intervention, suggesting different degrees of sensitivity to visual information at this critical layer.

These observations motivate us to perform visual knowledge distillation at the model's PVRL, where visual information has a particularly strong influence on response predictions. Specifically, DVTR-KD employs a reconstructor to recover the teacher's dense visual representations from the student's compressed visual states. The reconstructed representations are then processed by the remaining teacher layers, providing teacher-guided supervision that encourages the student to preserve visual information essential for subsequent language generation. We also align the student's native attention outputs with those of the original teacher at the PVRL. Together with response distillation and final-layer visual distillation, these objectives supervise both the reconstructed visual states and the student's own use of multimodal context.

Extensive experiments demonstrate that DVTR-KD consistently improves the performance of compressed MLLMs across different model scales and compression ratios.

In conclusion, our contributions are as follows:
\begin{itemize}
    \item We propose DVTR-KD, which reconstructs dense visual states from compressed student tokens to enable distillation between models with different visual sequence lengths.
    \item We use layer-wise visual-value interventions to select an intermediate reconstruction layer. We supervise reconstruction at the selected layer using response predictions from the remaining teacher layers.
    \item We evaluate DVTR-KD on various benchmarks. The experimental results demonstrate that across the evaluated settings, DVTR-KD achieves state-of-the-art (SOTA) performance, without adding modules or computation at inference time.
\end{itemize}

\section{Related work}
\label{sec:related}

\subsection{Multimodal Large Language Models}
Following the remarkable success of large language models (LLMs) in natural language processing \cite{achiam2023gpt,touvron2023llamaopenefficientfoundation}, substantial efforts have been devoted to extending their capabilities to visual understanding. Early MLLMs typically connect a pretrained visual encoder with a pretrained LLM through lightweight cross-modal modules. BLIP-2 \cite{li2023blip} introduces a Querying Transformer (Q-Former) to bridge frozen image encoders and LLMs, while LLaVA \cite{llava} adopts a simpler architecture that projects visual features into the LLM embedding space and further performs visual instruction tuning. Building upon the LLaVA architecture, subsequent models substantially improve visual resolution and task coverage. LLaVA-NeXT \cite{liu2024llavanext} introduces higher-resolution visual inputs to enhance fine-grained perception and OCR capabilities, while LLaVA-OneVision \cite{llava_onevision} further unifies single-image, multi-image, and video understanding within a single model. More recent MLLMs support flexible-resolution visual inputs: Qwen3-VL \cite{bai2025qwen3} dynamically represents images of different resolutions with different numbers of visual tokens, while Kimi-VL \cite{kimi_vl} employs native-resolution visual encoding to preserve fine-grained image information. These advances have greatly enhanced the perceptual and reasoning capabilities of MLLMs, but processing high-resolution visual inputs also introduces increasingly long visual sequences and substantial computational overhead.

\subsection{Distillation for MLLMs}

Knowledge distillation has been extensively studied for transferring the capabilities of large language models to students. MiniLLM \cite{ICLR2024_8ac015d4} introduces an on-policy method and performs distillation on student-generated sequences instead of ground-truth responses, while DistiLLM \cite{distillm} develops a skewed-KL objective together with an adaptive off-policy strategy for more efficient LLM distillation.

Recent studies further extend knowledge distillation to multimodal language models. LLaVA-KD \cite{cai2025llava} provides a general MLLM distillation framework that combines response distillation with visual representation. Align-KD \cite{align-kd} focuses on cross-modal knowledge transfer by distilling first-layer vision--language matching. More recently, Switch-KD \cite{switch_kd} instead feeds student visual representations into the teacher's language pathway, enabling multimodal supervision within a shared text-probability space. DVTR-KD instead reconstructs the teacher's intermediate visual states from the student's visual representations and optimizes this reconstruction using response-level supervision.

Most existing methods assume compatible visual representations between teacher and student and do not explicitly address the unequal visual sequence lengths introduced by token compression. EM-KD \cite{em-kd} directly studies this problem and employs Hungarian matching to establish one-to-one correspondences between teacher and student visual tokens, followed by visual semantic and vision--language affinity distillation. However, when the student is highly compressed, only the matched subset of teacher visual tokens directly participates in token-level supervision. In contrast, DVTR-KD reconstructs a dense visual representation from compressed student tokens and aligns it with the complete teacher sequence, allowing all teacher visual positions to contribute to direct visual supervision.
\section{Method}
\label{sec:method}

\begin{figure*}[t]
    \centering
    \includegraphics[width=\textwidth]{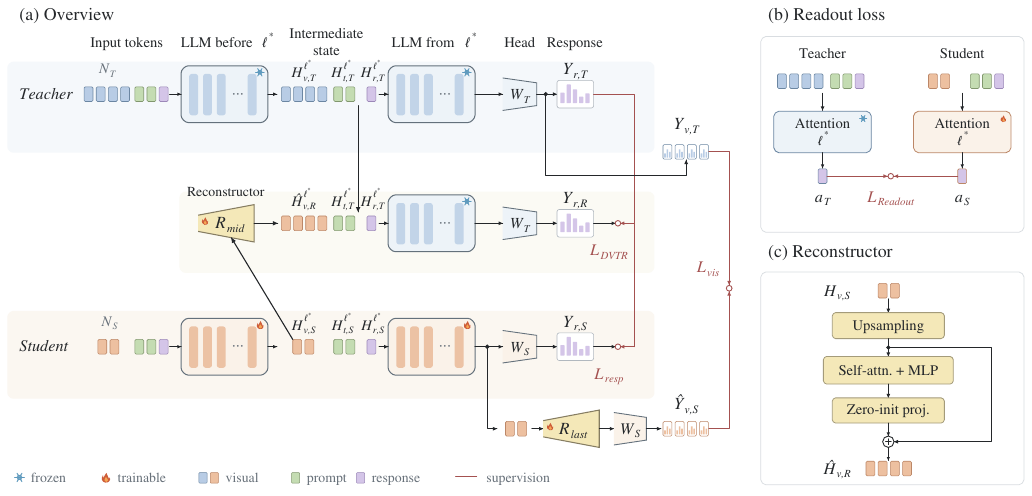}
    \caption{Overview of DVTR-KD. (a) The intermediate reconstructor $R_{\mathrm{mid}}$ reconstructs student visual states in the teacher's dense layout at PVRL, substituting teacher's visual states; this reconstruction is supervised through response predictions produced by the teacher from the PVRL onward. Native response distillation and an independent final-layer reconstructor $R_{\mathrm{last}}$ provide additional supervision. (b) Readout alignment uses attention outputs from the original teacher and the student. (c) An illustration to the reconstructor.}
    \label{fig:dvtr_overview}
\end{figure*}

\subsection{Composition of Distilled MLLM}
\label{sec:composition}
\paragraph{Visual Encoder.}
Both teacher and student use the same pretrained Vision Transformer (ViT)~\cite{dosovitskiy2021an} as their visual backbone.
Given an image $X_v \in \mathbb{R}^{H \times W \times 3}$, each model pads it to $H' \times W'$, with both dimensions chosen to accommodate the patch size $S_p$ and its visual-token compression configuration.
The padded image is divided into $N_p = H'W'/S_p^2$ non-overlapping patches, from which the ViT extracts spatial visual features.

\paragraph{Visual Projector and Compressor.}
The visual projector maps visual features into the LLM embedding space.
The student reduces the visual-token count through spatial compression within the ViT and/or in the projector.
We denote the overall spatial downsampling factor by $\rho$, so that $\rho^2$ is the nominal visual-token compression factor.
Let $T$ and $S$ denote the teacher and student models, respectively, and let $N_{v,T}$ and $N_{v,S}$ denote their visual-token counts at the LLM input, with $N_{v,S}<N_{v,T}$.

\paragraph{Large Language Model (LLM).}
The LLM processes visual and prompt embeddings $H=[H_v,H_t]$. We denote the output logits by $Z=[Z_v,Z_t,Z_r]$ and the corresponding vocabulary probability distributions by $Y=[Y_v,Y_t,Y_r]$, where $v$, $t$, and $r$ indicate visual, prompt, and response positions, respectively. For each position $i$, the probability distribution is computed as
\begin{equation}
    \mathbf{y}_i
    = \frac{\exp(\mathbf{z}_i/\tau)}
    {\left\|\exp(\mathbf{z}_i/\tau)\right\|_1},
\label{eq:probabilities}
\end{equation}
where $\tau$ is the temperature, $K$ is the vocabulary size, and $\mathbf{z}_i,\mathbf{y}_i \in \mathbb{R}^{K}$.

\subsection{Training Scheme}
\label{sec:schema}

Before PVRL analysis and distillation, we prepare teacher and student MLLMs through pretraining and SFT.

\paragraph{Stage 1: Pretraining.}
We freeze the pretrained visual encoder and LLM, and train the visual projector together with any learnable student compressor on image-caption pairs. This stage aligns visual representations with the LLM embedding space. For powerful teachers we further introduce an optional stage following LLaVA-OneVision~\cite{llava_onevision}, unfreezing the teacher LLM and jointly training it with the projector on detailed image descriptions.

\paragraph{Stage 2: SFT.}
We fine-tune both teacher and student on multimodal instruction-following data, updating the LLM, projector, and any learnable compressor while keeping the visual encoder frozen. The resulting SFT models are used for PVRL analysis, and each distillation run starts from the corresponding SFT checkpoint.

\paragraph{Supervision objective.}
All preparation stages minimize the autoregressive cross-entropy over response tokens. For model $M\in\{T,S\}$, let $Y_{r,M}$ denote its predicted response distributions and $Y_{\mathrm{gt}}$ the corresponding ground-truth distributions, where each $\mathbf{y}_{\mathrm{gt},j}$ is a one-hot probability vector.
\begin{equation}
\mathcal{C} (Y_{r,M},Y_{\mathrm{gt}})
=-\frac{1}{N_r}\sum_{j=1}^{N_r}
\left\langle \mathbf{y}_{\mathrm{gt},j}, \log \mathbf{y}_{r,M,j} \right\rangle.
\label{eq:ce}
\end{equation}
Here, $N_r$ is the number of response tokens, $\mathcal{C}$ represents cross entropy, $\langle\cdot,\cdot\rangle$ denotes the inner product, and the logarithm is applied element-wise. Each predicted distribution is conditioned on the image, prompt, and preceding ground-truth response tokens.

\subsection{Principal Visual Readout Layer}
\label{sec:pvrl}

We identify the principal visual readout layer (PVRL) by measuring how suppressing visual information at each LLM layer affects response prediction.
For a given LLM $M$ with $L$ transformer layers, we intervene on one layer at a time.
At layer $\ell$, we set the value vectors at visual-token positions to zero, leaving the queries, keys, text-token values, and attention normalization unchanged.
This removes the direct contribution of visual values to that layer's attention output while preserving its residual connection. The remaining LLM computations proceed normally. We denote the resulting model by $M_{\ell}^{V=0}$.

We evaluate the original and intervened models on the same fixed diagnostic subset sampled from the SFT dataset. Let $d$ denote the diagnostic set and $\mathrm{NLL}_i(M)$ denote the mean negative log-likelihood of the response tokens in example $i$ under model $M$. The sensitivity score of layer $\ell$ is
\begin{equation}
\Delta \mathrm{NLL}_{\ell}
=
\frac{1}{|d|}
\sum_{i \in d}
\left[
\mathrm{NLL}_i(M_{\ell}^{V=0})
-
\mathrm{NLL}_i(M)
\right].
\label{eq:pvrl_score}
\end{equation}
Thus, response tokens are averaged within each example. A larger positive $\Delta\mathrm{NLL}_\ell$ means that zeroing visual-token values at layer $\ell$ causes a greater reduction in the likelihood assigned to ground-truth responses, indicating greater sensitivity to the removal of direct visual-value contributions at that layer. We select the layer with the largest sensitivity score as the PVRL:
\begin{equation}
\ell^*
=
\mathop{\arg\max}_{\ell \in \{1,\ldots,L\}}
\Delta \mathrm{NLL}_{\ell}.
\label{eq:pvrl}
\end{equation}
Note that this procedure is applied independently to each model under analysis, with its parameters held fixed.

\subsection{DVTR-KD}
\label{sec:dvtr}

\paragraph{Reconstructor.}
To bridge the token-count mismatch, we upsample the student's coarse visual grid to the teacher grid and learn a residual correction.
Because teacher and student use different padding, we map teacher-token centers to the student grid using image coordinates.
The operator $B$ bilinearly interpolates neighboring student states at these locations, downweighting cells that overlap padding.
Given student visual states $H_{v,S}$, we compute
\begin{equation}
\begin{aligned}
U &= B(H_{v,S}), \\
\hat H_{v,R} &= R(H_{v,S})
= U+\varphi_{\theta}(U+\mathrm{pos}),
\end{aligned}
\label{eq:reconstructor}
\end{equation}
where $\mathrm{pos}$ contains 2D positional embeddings.
$\hat H_{v,R}\in\mathbb{R}^{N_{v,T}\times D}$ contains $N_{v,T}$ reconstructed visual states, where $D$ is the LLM hidden dimension.
The refinement module $\varphi_{\theta}$ consists of one bidirectional transformer layer followed by a zero-initialized output projection, so reconstruction initially equals the upsampled states $U$.
We use independent reconstructors $R_{\mathrm{last}}$ and $R_{\mathrm{mid}}$ for final-layer and intermediate-layer supervision, respectively.
Implementation details are provided in supplementary materials.

\paragraph{Visual Semantic Alignment.}
Aligning vocabulary distributions at visual-token positions has proved effective in MLLM distillation, as demonstrated by the visual component of LLaVA-KD~\cite{cai2025llava} and EM-KD~\cite{em-kd}. However, the token-wise alignment cannot be directly applied when teacher and student have unequal token counts. EM-KD addresses this mismatch through one-to-one token matching, but directly supervises at most $N_{v,S}$ teacher positions, leaving the remaining $(N_{v,T}-N_{v,S})$ tokens without a matched target. We instead reconstruct a dense student representation and apply visual semantic supervision across the teacher grid.

Let $H_{v,S}$ denote the student visual states after the final LLM layer.
We reconstruct $\hat{H}_{v,S}=R_{\mathrm{last}}(H_{v,S})$ on the teacher grid, then apply the student LLM head and the softmax in \cref{eq:probabilities} to obtain $\hat{Y}_{v,S} = \{\hat{\mathbf{y}}_{v,S,j}\}_{j=1}^{N_{v,T}}$. The reconstructed student distributions $\hat{Y}_{v,S}$ are aligned with the teacher distributions $Y_{v,T}$ at corresponding grid positions. We denote $\mathcal{D}$ as the average of KL divergences scaled by $\tau$ denoted in Eq~\ref{eq:probabilities} over corresponding token positions in two collections of distributions. The visual semantic alignment loss is
\begin{equation}
\begin{aligned}
\mathcal L_{\mathrm{vis}}
&=\mathcal{D} (Y_{v,T}\,\Vert\,\hat{Y}_{v,S}) \\
&= \frac{1}{N_{v,T}} \sum^{N_{v,T}}_{j=1} \tau ^{2} \operatorname{KL}(\mathbf{y}_{v,T,j} \Vert \hat{\mathbf{y}}_{v,S,j}) 
\label{eq:visual}
\end{aligned}
\end{equation}

\paragraph{Reconstruction at PVRL.}
Align-KD~\cite{align-kd} demonstrates the benefits of distilling cross-modal alignment at the first LLM layer, highlighting the value of supervision beyond final-layer outputs. We further consider where intermediate distillation can be more effective. Our layer-wise diagnostic motivates choosing the PVRL, where suppressing visual values causes the largest increase in response NLL. We therefore reconstruct dense visual states at the input to this layer. Under strong visual-token compression, however, recovering every component of the teacher's hidden states is challenging and may be unnecessary for answering the question. We instead supervise reconstruction through the teacher's response predictions rather than using token-wise MSE or KL loss, encouraging the student to preserve visual information that supports answering.

Let $H_M^{\ell^*}$ denote the states entering the selected layer.
We reconstruct the student visual states as $\hat H_{v,R}=R_{\mathrm{mid}}(H_{v,S}^{\ell^*})$ and substitute them for all visual positions in $H_{v,T}^{\ell^*}$, forming $\widetilde H_{T}^{\ell^*} = [\hat H_{v,R}, H_{t,T}^{\ell^*}, H_{r,T}^{\ell^*} ]$. The original teacher prompt states $H_{t,T}^{\ell^*}$ and response states $H_{r,T}^{\ell^*}$ are retained. The resulting sequence is processed by the teacher's subsequent layers:
\begin{equation}
Y_R = \phi_T^{\ell^*:L}(\widetilde H_T^{\ell^*}),
\label{eq:replay}
\end{equation}
where $\phi_T^{\ell^*:L}$ contains the remaining teacher layers and LLM head, followed by the normalization in \cref{eq:probabilities}.
Let $Y_{r,R}$ denote the response distributions in $Y_R$. Using the CE objective defined in \cref{eq:ce}, we combine ground-truth supervision with response distillation from the original teacher:
\begin{equation}
\mathcal L_{\mathrm{DVTR}}=\lambda\mathcal{C}(Y_{r,R},Y_{\mathrm{gt}})+(1-\lambda)\mathcal{D}(Y_{r,T}\Vert Y_{r,R}).
\label{eq:functional}
\end{equation}
This intermediate reconstruction is optimized for its contribution to answering, without directly regressing each reconstructed visual state to its teacher counterpart.

\paragraph{Attention-Readout Alignment.}
\label{sec:readout}

As an auxiliary objective, we introduce attention-readout alignment to guide the student in learning what information the teacher retrieves at the PVRL. Specifically, we align the native student's attention outputs with those of the teacher at response positions, encouraging the student to extract similar information. Let $a_{r,M}^{\ell^*}\in\mathbb{R}^{N_{r} \times D}$ denote the attention output after output projection at layer $\ell^*$ for the $N_r$ supervised response positions. The readout loss is defined as

\begin{equation}
\mathcal L_{\mathrm{readout}}
=\mathrm{MSE} (a_{r,S}^{\ell^*} , a_{r,T}^{\ell^*} )
\label{eq:readout}
\end{equation}

\paragraph{Native Response Supervision.}
We combine ground-truth supervision with response distillation from the original teacher for the native student:
\begin{equation}
\mathcal L_{\mathrm{resp}}=\lambda\mathcal{C}(Y_{r,S},Y_{\mathrm{gt}})+({1-\lambda})\mathcal{D}(Y_{r,T}\Vert Y_{r,S}).
\label{eq:student}
\end{equation}

\subsection{Complete Training Objective}
\label{sec:objective}

The complete training objective combines native response supervision with intermediate reconstruction, attention-readout alignment, and final-layer visual semantic alignment:
\begin{equation}
\begin{aligned}
\mathcal L={}&\mathcal L_{\mathrm{resp}}+\lambda_{\mathrm{D}}\mathcal L_{\mathrm{DVTR}}
+\lambda_{\mathrm{r}}\mathcal L_{\mathrm{readout}}+\lambda_{\mathrm{v}}\mathcal L_{\mathrm{vis}}.
\end{aligned}
\label{eq:total}
\end{equation}
\section{Experiments}
\label{sec:experiments}

\begin{table*}[t]
    \centering
    \caption{Training data for model preparation and knowledge distillation.
    ``--'' indicates a skipped stage; ``Frozen'' indicates that the teacher is used without parameter updates during KD.}
    \label{tab:training_data}
    \small
    \setlength{\tabcolsep}{6pt}
    \renewcommand{\arraystretch}{1.2}
    \begin{tabular}{@{}lcccc@{}}
        \toprule
        Model
        & Stage 1 
        & Stage 1.5 
        & SFT 
        & KD \\
        \midrule
        Student
        & LLaVA-PreTrain-558K
        & --
        & LLaVA-NeXT-779K
        & LLaVA-NeXT-779K \\
        Same-data teacher
        & LLaVA-PreTrain-558K
        & --
        & LLaVA-NeXT-779K
        & Frozen \\
        Data-rich teacher
        & LLaVA-PreTrain-558K
        & LLaVA-ReCap-558K
        & \shortstack{LLaVA-OneVision-SI-3.3M}
        & Frozen \\
        \bottomrule
    \end{tabular}
\end{table*}

\begin{table*}[t]
\centering
\caption{Distillation results across visual-token compression ratios, teacher data regimes and teacher scales. All methods are re-implemented for fair comparison and start from the corresponding SFT model. $\rho^2$ denotes the nominal visual-token compression factor. LLaVA-KD$^{*}$ means we use our reconstruction method to align different visual-token counts after last block. MME scores are divided by 28 when calculating the average scores. $Avg_{11}$ is the mean over all 11 benchmarks.}
\label{tab:main_results}
\setlength{\tabcolsep}{2.5pt}
\renewcommand{\arraystretch}{1.12}
\footnotesize
\resizebox{0.85\textwidth}{!}{%
\begin{tabular}{@{}lc*{12}{c}@{}}
\toprule
& & \multicolumn{3}{c}{General} & \multicolumn{4}{c}{Knowledge} & \multicolumn{4}{c}{OCR} & \\
\cmidrule(lr){3-5}\cmidrule(lr){6-9}\cmidrule(lr){10-13}
Method & $\rho^2$ & GQA & MME & RWQA & SQA & AI2D & MMMU & MMStar & ChartQA & DocVQA & TextVQA & OCRBench & $Avg_{11}$ \\
\midrule
\multicolumn{14}{c}{\textit{Qwen3-0.6B teacher trained on 4.4M samples $\rightarrow$ Qwen3-0.6B compressed students}} \\
\midrule
Teacher & $1\times$ & 56.6 & 1862.9 & 61.7 & 88.2 & 73.6 & 37.0 & 50.2 & 78.8 & 87.5 & 77.3 & 74.3 & 68.3 \\
\midrule
SFT  & $4\times$ & 57.8 & 1450.2 & 53.6 & 59.8 & 59.9 & \underline{32.1} & 38.3 & 64.6 & 67.7 & 63.1 & 49.8 & 54.4 \\
MiniLLM & $4\times$ & \underline{58.3} & 1616.3 & 56.1 & 61.3 & 60.8 & \underline{32.1} & 38.6 & 64.6 & 66.9 & 65.1 & 52.2 & 55.8 \\
LLaVA-KD$^{*}$ & $4\times$ & \textbf{59.7} & \underline{1636.3} & \underline{58.0} & \underline{62.6} & \underline{64.9} & \textbf{32.3} & \underline{40.5} & \underline{70.6} & \underline{74.4} & \underline{68.0} & \underline{56.3} & \underline{58.7} \\
EM-KD & $4\times$ & \underline{58.3} & 1461.6 & 55.0 & 62.4 & 63.1 & 30.1 & 39.9 & 69.0 & 73.7 & 66.3 & 53.8 & 56.7 \\
\textbf{Ours} & $4\times$ & \textbf{59.7} & \textbf{1648.3} & \textbf{59.2} & \textbf{65.4} & \textbf{67.4} & 30.6 & \textbf{40.8} & \textbf{73.0} & \textbf{77.7} & \textbf{70.0} & \textbf{56.7} & \textbf{59.9} \\
\midrule
SFT  & $16\times$ & 56.0 & 1408.3 & 53.7 & 61.6 & 60.7 & 32.9 & 38.1 & 52.2 & 45.4 & 51.0 & 32.1 & 48.5 \\
MiniLLM & $16\times$ & 56.6 & \underline{1528.7} & \textbf{56.7} & 62.0 & 61.0 & \textbf{35.6} & 37.7 & 53.4 & 46.1 & 52.9 & \underline{35.0} & 50.1 \\
LLaVA-KD$^{*}$ & $16\times$ & \underline{57.9} & 1347.2 & 54.6 & \underline{62.4} & \textbf{65.3} & 33.0 & 37.5 & 58.6 & \underline{51.5} & \underline{53.9} & 34.0 & \underline{50.6} \\
EM-KD & $16\times$ & 56.7 & 1379.8 & 52.4 & 60.4 & \underline{62.8} & 32.4 & \underline{38.6} & \underline{59.4} & 51.1 & 53.4 & 33.1 & 50.0 \\
\textbf{Ours} & $16\times$ & \textbf{58.0} & \textbf{1549.1} & \underline{56.5} & \textbf{64.9} & \textbf{65.3} & \underline{33.8} & \textbf{39.7} & \textbf{62.6} & \textbf{55.2} & \textbf{56.8} & \textbf{35.1} & \textbf{53.0} \\
\midrule
\multicolumn{14}{c}{\textit{Qwen3-0.6B teacher trained on 1.3M samples $\rightarrow$ Qwen3-0.6B compressed students}} \\
\midrule
Teacher & $1\times$ & 58.9 & 1567.2 & 56.6 & 62.3 & 64.9 & 32.8 & 40.7 & 71.6 & 74.8 & 67.9 & 59.7 & 58.7 \\
\midrule
SFT  & $4\times$ & 57.8 & 1450.2 & 53.6 & 59.8 & 59.9 & \textbf{32.1} & 38.3 & 64.6 & 67.7 & 63.1 & 49.8 & 54.4 \\
MiniLLM & $4\times$ & 58.2 & 1458.2 & 53.6 & \underline{61.1} & 60.3 & \underline{31.8} & 37.4 & 64.0 & 67.6 & 64.8 & 52.1 & 54.8 \\
LLaVA-KD$^{*}$ & $4\times$ & \underline{59.0} & \textbf{1553.6} & \underline{53.9} & \underline{61.1} & \underline{63.5} & 30.7 & \underline{39.5} & \underline{69.0} & \underline{71.8} & \underline{65.8} & 53.4 & \underline{56.6} \\
EM-KD & $4\times$ & 58.4 & 1451.8 & \underline{53.9} & 59.6 & 62.1 & 30.1 & 39.3 & 68.2 & 71.4 & 64.4 & \underline{54.5} & 55.8 \\
\textbf{Ours} & $4\times$ & \textbf{59.3} & \underline{1523.7} & \textbf{56.3} & \textbf{61.6} & \textbf{64.1} & 30.6 & \textbf{39.6} & \textbf{70.8} & \textbf{73.5} & \textbf{66.9} & \textbf{55.3} & \textbf{57.5} \\
\midrule
\multicolumn{14}{c}{\textit{Qwen3-4B teacher trained on 4.4M samples $\rightarrow$ Qwen3-4B compressed students}} \\
\midrule
Teacher & $1\times$ & 61.4 & 2163.3 & 67.3 & 96.8 & 82.2 & 48.8 & 63.9 & 84.2 & 93.0 & 81.4 & 78.7 & 75.9 \\
\midrule
SFT  & $4\times$ & 61.3 & 1896.4 & 61.4 & 78.1 & 76.0 & \textbf{44.0} & \underline{49.3} & 76.4 & 80.9 & 71.2 & 63.1 & 66.3 \\
MiniLLM & $4\times$ & 61.5 & \underline{1961.3} & 61.4 & 78.9 & \underline{76.8} & \underline{43.2} & 47.9 & 72.9 & 81.2 & 74.4 & 63.2 & 66.5 \\
LLaVA-KD$^{*}$ & $4\times$ & \underline{62.5} & 1880.2 & \textbf{64.1} & 78.3 & \underline{76.8} & 42.4 & 48.9 & \underline{78.3} & \underline{84.5} & \underline{75.4} & \textbf{68.1} & \underline{67.9} \\
EM-KD & $4\times$ & 61.8 & 1954.0 & 63.0 & \underline{79.3} & 76.5 & 41.7 & 47.6 & 77.1 & 83.0 & 73.8 & 65.4 & 67.2 \\
\textbf{Ours} & $4\times$ & \textbf{62.8} & \textbf{2008.6} & \underline{63.7} & \textbf{80.9} & \textbf{77.8} & \textbf{44.0} & \textbf{49.9} & \textbf{80.9} & \textbf{87.4} & \textbf{77.5} & \underline{67.9} & \textbf{69.5} \\
\bottomrule
\end{tabular}%
}
\end{table*}

\subsection{Experimental Settings}
\label{sec:setting}

\paragraph{Implementation Details.}  We adopt two variants of Qwen3 \cite{qwen3}, Qwen3-0.6B and Qwen3-4B, as the language backbones to construct two teacher-student configurations. Within each configuration, the teacher and student share the same LLM architecture, while differing in their visual token compression strategies. We primarily conduct experiments with Qwen3-0.6B and further extend our evaluation to Qwen3-4B to validate the generality of our method across different model scales. Both teacher and student models are initialized from the same officially released text-only LLM checkpoints, without prior multimodal fine-tuning. For visual feature extraction, we employ a pretrained MoonViT-SO-400M \cite{kimi_vl} as the frozen visual encoder, which supports native-resolution image processing and produces fine-grained visual representations. 

For the uncompressed teacher models, we employ a two-layer MLP projector to map visual representations into the LLM embedding space without reducing the number of visual tokens. For compressed models, we investigate two visual-token compression ratios, $\rho^2 \in \{4, 16\}$. For the $4\times$ compressed models, we adopt a pixel-unshuffle projector \cite{dong2024internlmxcomposer4khd}, which concatenates each $2\times2$ group of spatially adjacent visual tokens along the channel dimension, followed by projection into the LLM embedding space. For the $16\times$ compressed models, inspired by LLaVA-UHD v4 \cite{llava_uhd}, we additionally insert a compressor after the 6-th ViT layer. This intra-ViT module employs local window attention to aggregate neighboring visual information, followed by $2\times2$ pixel-unshuffle and feature fusion to reduce the token count by a factor of four. Combined with the final pixel-unshuffle projector, this achieves an overall $16\times$ compression ratio. To accommodate different compression ratios, we adopt ratio-dependent image padding strategies. Specifically, the image height and width are padded to multiples of 14 pixels for the teacher, 28 pixels for students with $\rho^{2}=4$, and 56 pixels for students with $\rho^{2}=16$. 

The experiments are conducted on $8\times$ NVIDIA A800 GPUs. The weights of loss functions are $\lambda=0.5$, $\lambda_{D}=2$, $\lambda_r=1$ and $\lambda_v=0.5$. Based on the PVRL analysis in Figure~\ref{fig:visual_readout_layers}, we use $\ell^*=12$ and $\ell^*=20$ for Qwen3-0.6B and Qwen3-4B, respectively.  More implementation details are shown in supplementary materials.

\paragraph{Training Datasets.}
As shown in Table~\ref{tab:training_data}, we compare distillation from \textbf{same-data teachers}, trained on the same data as the students, and \textbf{data-rich teachers}, trained with additional data.
Before distillation, we prepare the models through the following stages.
The ViT remains frozen throughout.

In \textbf{Stage 1 (pretraining)}, all models use 558K image-caption pairs~\cite{llava} to align visual representations with the LLM embedding space.
We train the projector and any learnable student compressor while freezing the LLM.
Only data-rich teachers undergo \textbf{Stage 1.5}, where we jointly train the LLM and projector on detailed image descriptions from LLaVA-ReCap-558K~\cite{liu2024llavanext} to improve visual understanding.

In \textbf{Stage 2 (visual instruction tuning)}, students and same-data teachers use the LLaVA-NeXT 779K instruction-tuning dataset~\cite{liu2024llavanext}.
Data-rich teachers use approximately 3.3M instruction-tuning samples following the LLaVA-OneVision single-image recipe~\cite{llava_onevision}.
This stage updates the LLM, projector, and any learnable student compressor.

For \textbf{knowledge distillation}, all methods use the LLaVA-NeXT 779K instruction-tuning dataset and initialize the student from its corresponding Stage-2 SFT checkpoint.

\paragraph{Baseline Methods.} To demonstrate the effectiveness of DVTR-KD, we make fair comparisons with several representative methods: MiniLLM~\cite{ICLR2024_8ac015d4}, LLaVA-KD~\cite{cai2025llava} and EM-KD~\cite{em-kd}. Note that LLaVA-KD cannot be directly applied to MLLMs with unbalanced visual tokens. We re-implement it with a reconstructor to align the number of visual tokens between teacher and student MLLMs, which is one of the key contributions in our work.

\paragraph{Benchmarks.}
We evaluate the effectiveness of our method on 11 benchmarks. They are categorized into three groups based on different areas. (1) General Question Answering: GQA~\cite{gqa}, MME~\cite{mme}, RealWorldQA (RWQA) \cite{xai2024grok15v}; (2) Comprehensive Knowledge Reasoning: ScienceQA-IMG (SQA)~\cite{scienceqa}, AI2D~\cite{ai2d}, MMMU~\cite{mmmu}, MMStar~\cite{mmstar}; (3) OCR and Chart Parsing: ChartQA~\cite{chartqa}, DocVQA~\cite{docvqa}, TextVQA~\cite{textvqa}, and OCRBench~\cite{ocrbench}. To compare the performance, we report the mean score across all 11 benchmarks, dividing MME scores by 28 before averaging.

\subsection{Main Results}
\label{sec:main_results}

As shown in Table~\ref{tab:main_results}, DVTR-KD consistently outperforms competing distillation methods across different experimental settings, achieving the highest average scores. 

With a same-data teacher, DVTR-KD reduces the average-score gap between the $4\times$ SFT student and its uncompressed teacher from 4.3 to 1.2 points. Using the data-rich teacher further improves the student from 57.5 to 59.9.

Specifically, comparing the same-data teacher and compressed students without distillation, we observe that aggressive visual token compression has a relatively limited impact on General and Knowledge tasks, but substantially degrades OCR performance. Specifically, the average non-OCR scores are 53.2, 50.5, and 50.5 for the uncompressed same-data teacher and students with compression ratios $\rho^{2}=4$ and $\rho^{2}=16$, respectively. In contrast, the corresponding mean OCR scores drop sharply from 68.5 to 61.3 and 45.2, highlighting the vulnerability of fine-grained visual perception to aggressive token compression. DVTR-KD effectively alleviates this degradation, achieving consistent improvements across OCR benchmarks. Compared with the strongest competing baseline, DVTR-KD improves $Avg_{11}$ by 0.9--1.6 points under $\rho^{2}=4$ and 2.4 points under $\rho^{2}=16$, demonstrating its effectiveness in transferring fine-grained visual knowledge to compressed MLLMs.

\subsection{Ablation Study}
\label{sec:ablation}

We conduct ablation studies to evaluate the effectiveness of each component in DVTR-KD. Unless otherwise specified, experiments are conducted on 0.6B students with $\rho^{2}=16$ and a data-rich teacher.

\paragraph{Key Components.}
We progressively introduce $\mathcal{L}_{\mathrm{vis}}$, $\mathcal{L}_{\mathrm{DVTR}}$, and $\mathcal{L}_{\mathrm{readout}}$ to the response distillation baseline $\mathcal{L}_{\mathrm{resp}}$. As shown in Table~\ref{tab:key_components}, each component contributes to performance improvements, increasing $Avg_{11}$ from 51.0 to 51.2, 52.6, and 53.0, respectively. The full objective achieves the best performance, demonstrating the effectiveness of combining these components.

\begin{table}[t]
  \centering
  \caption{Ablation study on contributions from key components in our method.}
  \label{tab:key_components}
  \small
  \setlength{\tabcolsep}{10pt}
  \renewcommand{\arraystretch}{1.25}
  \begin{tabular}{@{}ccccr@{}}
    \toprule
    $\mathcal{L}_{\mathrm{resp}}$ &
    $\mathcal{L}_{\mathrm{vis}}$ &
    $\mathcal{L}_{\mathrm{DVTR}}$ &
    $\mathcal{L}_{\mathrm{readout}}$ &
    $Avg_{11}$ \\
    \midrule
    \checkmark & --         & --         & --         & 51.0 \\
    \checkmark & \checkmark & --         & --         & 51.2   \\
    \checkmark & \checkmark & \checkmark & --         & 52.6   \\
    \midrule
    \checkmark & \checkmark & \checkmark & \checkmark & \textbf{53.0} \\
    \bottomrule
  \end{tabular}
\end{table}

\paragraph{Supervision for Reconstruction at PVRL.}
To optimize $\mathcal{L}_{\mathrm{DVTR}}$, we replace the teacher's visual states at the PVRL with student-reconstructed representations and apply KL divergence and cross-entropy for response-level supervision. We compare this approach with direct feature alignment using MSE loss. As shown in Table~\ref{tab:dvtr_supervision}, compared with response-level supervision, MSE slightly increases $Avg_{11}$ by 0.3 points under $\rho^2=4$ but significantly decreases it by 1.1 points under $\rho^2=16$. These results suggest that strict feature matching becomes less effective under aggressive compression, while response-guided supervision is more robust, performing well on both compression ratios.

\begin{table}[t]
  \centering
  \caption{Ablation study on supervision type for reconstruction at PVRL.}
  \label{tab:dvtr_supervision}
  \small
  \setlength{\tabcolsep}{10pt}
  \renewcommand{\arraystretch}{1.25}
  \begin{tabular}{@{}ccr@{}}
    \toprule
    $\rho^{2}$ & type & $Avg_{11}$ \\ 
    \midrule
    $4\times$ & MSE & 60.2 \\ 
    $4\times$ & response & 59.9 \\ 
    \midrule 
    $16\times$ & MSE & 51.9 \\ 
    $16\times$ & response & 53.0 \\
    \bottomrule
  \end{tabular}
\end{table}

\paragraph{Reconstruction at Different Layers.}
Align-KD~\cite{align-kd} introduces first-layer cross-modal attention distillation, demonstrating the potential of supervision beyond final-layer outputs. Although it cannot be directly applied to mismatched visual sequences, we investigate reconstruction at the first layer ($\ell=1$) as well as two neighboring layers ($\ell=11,13$) of the PVRL ($\ell^*=12$). As shown in Table~\ref{tab:dvtr_layers}, reconstruction at the PVRL achieves the highest overall score, supporting the effectiveness of our layer selection strategy.

\begin{table}[t]
  \centering
  \caption{Reconstruction at different layers. $Avg_{G}$, $Avg_{K}$, and $Avg_{O}$ denote the mean scores on the General, Knowledge, and OCR benchmarks, respectively.}
  \label{tab:dvtr_layers}
  \small
  \setlength{\tabcolsep}{10pt}
  \renewcommand{\arraystretch}{1.25}
  \begin{tabular}{@{}ccccr@{}}
    \toprule
 $\ell$ & $Avg_{G}$ & $Avg_{K}$ & $Avg_{O}$ & $Avg_{11}$ \\ 
 \midrule
 1 & 55.5 & 49.7 & 51.6 & 52.0 \\
 11 & 56.6 & 50.8 & 51.6 & 52.7 \\
 13 & 55.9 & 50.5 & 52.3 & 52.6 \\ 
 \midrule
 12 ($\ell^{*}$) & 56.6 & 51.0 & 52.4 & 53.0 \\
    \bottomrule
  \end{tabular}
\end{table}
\section{Conclusion}
\label{sec:conclusion}

This paper proposes DVTR-KD, a novel knowledge distillation framework for transferring visual knowledge from a teacher MLLM to a token-compressed student. To address the mismatch in visual sequence lengths, DVTR-KD reconstructs dense visual representations from compressed student tokens, enabling token-wise alignment with the complete teacher visual sequence. We further introduce intermediate-layer distillation at PVRL, identified as the layer most sensitive to visual information removal. Specifically, we replace the teacher's visual states at the input to PVRL with student-reconstructed representations and supervise them through the remaining teacher layers using response-level objectives. Comprehensive experiments demonstrate the effectiveness of DVTR-KD.
{
    \small
    \bibliographystyle{ieeenat_fullname}
    \bibliography{main}

\begin{thebibliography}{40}
\providecommand{\natexlab}[1]{#1}
\providecommand{\url}[1]{\texttt{#1}}
\expandafter\ifx\csname urlstyle\endcsname\relax
  \providecommand{\doi}[1]{doi: #1}\else
  \providecommand{\doi}{doi: \begingroup \urlstyle{rm}\Url}\fi

\bibitem[Antol et~al.(2015)Antol, Agrawal, Lu, Mitchell, Batra, Zitnick, and Parikh]{antol2015vqa}
Stanislaw Antol, Aishwarya Agrawal, Jiasen Lu, Margaret Mitchell, Dhruv Batra, C~Lawrence Zitnick, and Devi Parikh.
\newblock Vqa: Visual question answering.
\newblock In \emph{Proceedings of the IEEE international conference on computer vision}, pages 2425--2433, 2015.

\bibitem[Bai et~al.(2023)Bai, Bai, Chu, Cui, Dang, Deng, Fan, Ge, Han, Huang, et~al.]{bai2023qwen}
Jinze Bai, Shuai Bai, Yunfei Chu, Zeyu Cui, Kai Dang, Xiaodong Deng, Yang Fan, Wenbin Ge, Yu Han, Fei Huang, et~al.
\newblock Qwen technical report.
\newblock \emph{arXiv preprint arXiv:2309.16609}, 2023.

\bibitem[Bai et~al.(2025)Bai, Cai, Chen, Chen, Chen, Cheng, Deng, Ding, Gao, Ge, et~al.]{bai2025qwen3}
Shuai Bai, Yuxuan Cai, Ruizhe Chen, Keqin Chen, Xionghui Chen, Zesen Cheng, Lianghao Deng, Wei Ding, Chang Gao, Chunjiang Ge, et~al.
\newblock Qwen3-vl technical report.
\newblock \emph{arXiv preprint arXiv:2511.21631}, 2025.

\bibitem[Cai et~al.(2025)Cai, Zhang, He, He, Tong, Gan, Wang, Xue, Liu, and Bai]{cai2025llava}
Yuxuan Cai, Jiangning Zhang, Haoyang He, Xinwei He, Ao Tong, Zhenye Gan, Chengjie Wang, Zhucun Xue, Yong Liu, and Xiang Bai.
\newblock {LLaVA-KD: A Framework of Distilling Multimodal Large Language Models}.
\newblock In \emph{{Proceedings of the IEEE/CVF International Conference on Computer Vision (ICCV)}}, pages 239--249, 2025.

\bibitem[Chen et~al.(2024)Chen, Li, Dong, Zhang, Zang, Chen, Duan, Wang, Qiao, Lin, and Zhao]{mmstar}
Lin Chen, Jinsong Li, Xiaoyi Dong, Pan Zhang, Yuhang Zang, Zehui Chen, Haodong Duan, Jiaqi Wang, Yu Qiao, Dahua Lin, and Feng Zhao.
\newblock {Are We on the Right Way for Evaluating Large Vision-Language Models?}
\newblock In \emph{Advances in Neural Information Processing Systems}, pages 27056--27087. Curran Associates, Inc., 2024.

\bibitem[{DeepSeek-AI} et~al.(2026){DeepSeek-AI}, Xu, Lin, Xue, Wang, Xu, Wu, Zhang, Lin, Dong, Ling, et~al.]{xu2026deepseek}
{DeepSeek-AI}, Anyi Xu, Bangcai Lin, Bing Xue, Bingxuan Wang, Bingzheng Xu, Bochao Wu, Bowei Zhang, Chaofan Lin, Chen Dong, Chenchen Ling, et~al.
\newblock Deepseek-v4: Towards highly efficient million-token context intelligence.
\newblock \emph{arXiv preprint arXiv:2606.19348}, 2026.

\bibitem[Dong et~al.(2024)Dong, Zhang, Zang, Cao, Wang, et~al.]{dong2024internlmxcomposer4khd}
Xiaoyi Dong, Pan Zhang, Yuhang Zang, Yuhang Cao, Bin Wang, et~al.
\newblock Internlm-xcomposer2-4khd: A pioneering large vision-language model handling resolutions from 336 pixels to 4k hd.
\newblock In \emph{Advances in Neural Information Processing Systems}, pages 42566--42592, 2024.

\bibitem[Dosovitskiy et~al.(2021)Dosovitskiy, Beyer, Kolesnikov, Weissenborn, Zhai, Unterthiner, Dehghani, Minderer, Heigold, Gelly, Uszkoreit, and Houlsby]{dosovitskiy2021an}
Alexey Dosovitskiy, Lucas Beyer, Alexander Kolesnikov, Dirk Weissenborn, Xiaohua Zhai, Thomas Unterthiner, Mostafa Dehghani, Matthias Minderer, Georg Heigold, Sylvain Gelly, Jakob Uszkoreit, and Neil Houlsby.
\newblock An image is worth 16x16 words: Transformers for image recognition at scale.
\newblock In \emph{International Conference on Learning Representations}, 2021.

\bibitem[Fang et~al.(2026)Fang, Qin, Wang, Ma, Yu, and Yao]{llava_uhd}
Kechen Fang, Yihua Qin, Chongyi Wang, Wenshuo Ma, Tianyu Yu, and Yuan Yao.
\newblock {LLaVA-UHD v4: What Makes Efficient Visual Encoding in MLLMs?}
\newblock arXiv:2605.08985, 2026.

\bibitem[Feng et~al.(2025)Feng, Li, Lin, and Chen]{align-kd}
Qianhan Feng, Wenshuo Li, Tong Lin, and Xinghao Chen.
\newblock {Align-KD: Distilling Cross-Modal Alignment Knowledge for Mobile Vision-Language Large Model Enhancement}.
\newblock In \emph{{Proceedings of the IEEE/CVF Conference on Computer Vision and Pattern Recognition (CVPR)}}, pages 4178--4188, 2025.

\bibitem[Feng et~al.(2026)Feng, Yang, Duan, Yang, and Wang]{em-kd}
Ze Feng, Sen Yang, Boqiang Duan, Wankou Yang, and Jingdong Wang.
\newblock {EM-KD: Distilling Efficient Multimodal Large Language Model with Unbalanced Vision Tokens}.
\newblock In \emph{{Proceedings of the AAAI Conference on Artificial Intelligence}}, pages 21111--21119. Association for the Advancement of Artificial Intelligence (AAAI), 2026.

\bibitem[Fu et~al.(2025)Fu, Chen, Shen, Qin, Zhang, Lin, Yang, Zheng, Li, Sun, Wu, Ji, Shan, and He]{mme}
Chaoyou Fu, Peixian Chen, Yunhang Shen, Yulei Qin, Mengdan Zhang, Xu Lin, Jinrui Yang, Xiawu Zheng, Ke Li, Xing Sun, Yunsheng Wu, Rongrong Ji, Caifeng Shan, and Ran He.
\newblock {MME: A Comprehensive Evaluation Benchmark for Multimodal Large Language Models}.
\newblock In \emph{Advances in Neural Information Processing Systems}. Curran Associates, Inc., 2025.

\bibitem[{GLM-5-Team} et~al.(2026){GLM-5-Team}, Zeng, Lv, Hou, Du, Zheng, Chen, Yin, Ge, Huang, Xie, et~al.]{zeng2026glm}
{GLM-5-Team}, Aohan Zeng, Xin Lv, Zhenyu Hou, Zhengxiao Du, Qinkai Zheng, Bin Chen, Da Yin, Chendi Ge, Chenghua Huang, Chengxing Xie, et~al.
\newblock Glm-5: from vibe coding to agentic engineering, 2026.

\bibitem[Goyal et~al.(2017)Goyal, Khot, Summers-Stay, Batra, and Parikh]{vqav2}
Yash Goyal, Tejas Khot, Douglas Summers-Stay, Dhruv Batra, and Devi Parikh.
\newblock Making the v in vqa matter: Elevating the role of image understanding in visual question answering.
\newblock In \emph{Proceedings of the IEEE Conference on Computer Vision and Pattern Recognition (CVPR)}, 2017.

\bibitem[Grattafiori et~al.(2024)Grattafiori, Dubey, Jauhri, Pandey, Kadian, Al-Dahle, Letman, Mathur, Schelten, Vaughan, et~al.]{grattafiori2024llama3}
Aaron Grattafiori, Abhimanyu Dubey, Abhinav Jauhri, Abhinav Pandey, Abhishek Kadian, Ahmad Al-Dahle, Aiesha Letman, Akhil Mathur, Alan Schelten, Alex Vaughan, et~al.
\newblock The llama 3 herd of models.
\newblock \emph{arXiv preprint arXiv:2407.21783}, 2024.

\bibitem[Gu et~al.(2024)Gu, Dong, Wei, and Huang]{ICLR2024_8ac015d4}
Yuxian Gu, Li Dong, Furu Wei, and Minlie Huang.
\newblock {MiniLLM: Knowledge Distillation of Large Language Models}.
\newblock In \emph{{International Conference on Learning Representations}}, pages 32694--32717, 2024.

\bibitem[Hudson and Manning(2019)]{gqa}
Drew~A. Hudson and Christopher~D. Manning.
\newblock {GQA: A New Dataset for Real-World Visual Reasoning and Compositional Question Answering}.
\newblock In \emph{Proceedings of the IEEE/CVF Conference on Computer Vision and Pattern Recognition (CVPR)}, pages 6700--6709, 2019.

\bibitem[Kembhavi et~al.(2016)Kembhavi, Salvato, Kolve, Seo, Hajishirzi, and Farhadi]{ai2d}
Aniruddha Kembhavi, Mike Salvato, Eric Kolve, Minjoon Seo, Hannaneh Hajishirzi, and Ali Farhadi.
\newblock {A Diagram Is Worth a Dozen Images}.
\newblock In \emph{Computer Vision -- ECCV 2016}, pages 235--251. Springer, 2016.

\bibitem[{Kimi Team} et~al.(2025){Kimi Team}, Du, Yin, Xing, Qu, Wang, Chen, Zhang, Du, Wei, et~al.]{kimi_vl}
{Kimi Team}, Angang Du, Bohong Yin, Bowei Xing, Bowen Qu, Bowen Wang, Cheng Chen, Chenlin Zhang, Chenzhuang Du, Chu Wei, et~al.
\newblock {Kimi-VL Technical Report}.
\newblock arXiv:2504.07491, 2025.

\bibitem[{Kimi Team} et~al.(2026){Kimi Team}, Bai, Bai, Bao, {M. C.}, Cai, Cai, Cao, Cao, Chai, Charles, et~al.]{kimik3}
{Kimi Team}, Tongtong Bai, Yifan Bai, Yiping Bao, {M. C.}, Jianfeng Cai, Xinyuan Cai, Peizhou Cao, Yuxuan Cao, Ziwei Chai, Y Charles, et~al.
\newblock Kimi k3: Open frontier intelligence.
\newblock \emph{arXiv preprint arXiv:2607.24653}, 2026.

\bibitem[Ko et~al.(2024)Ko, Kim, Chen, and Yun]{distillm}
Jongwoo Ko, Sungnyun Kim, Tianyi Chen, and Se-Young Yun.
\newblock {D}isti{LLM}: Towards streamlined distillation for large language models.
\newblock In \emph{Proceedings of the 41st International Conference on Machine Learning}, pages 24872--24895. PMLR, 2024.

\bibitem[Li et~al.(2025)Li, Zhang, Guo, Zhang, Li, Zhang, Zhang, Zhang, Li, Liu, and Li]{llava_onevision}
Bo Li, Yuanhan Zhang, Dong Guo, Renrui Zhang, Feng Li, Hao Zhang, Kaichen Zhang, Peiyuan Zhang, Yanwei Li, Ziwei Liu, and Chunyuan Li.
\newblock {LLaVA-OneVision: Easy Visual Task Transfer}.
\newblock \emph{Transactions on Machine Learning Research}, 2025.

\bibitem[Li et~al.(2023)Li, Li, Savarese, and Hoi]{li2023blip}
Junnan Li, Dongxu Li, Silvio Savarese, and Steven Hoi.
\newblock Blip-2: Bootstrapping language-image pre-training with frozen image encoders and large language models.
\newblock In \emph{International conference on machine learning}, pages 19730--19742. PmLR, 2023.

\bibitem[Liu et~al.(2023)Liu, Li, Wu, and Lee]{llava}
Haotian Liu, Chunyuan Li, Qingyang Wu, and Yong~Jae Lee.
\newblock {Visual Instruction Tuning}.
\newblock In \emph{{Advances in Neural Information Processing Systems}}, pages 34892--34916. Curran Associates, Inc., 2023.

\bibitem[Liu et~al.(2024{\natexlab{a}})Liu, Li, Li, and Lee]{llava15}
Haotian Liu, Chunyuan Li, Yuheng Li, and Yong~Jae Lee.
\newblock {Improved Baselines with Visual Instruction Tuning}.
\newblock In \emph{{Proceedings of the IEEE/CVF Conference on Computer Vision and Pattern Recognition (CVPR)}}, pages 26296--26306, 2024{\natexlab{a}}.

\bibitem[Liu et~al.(2024{\natexlab{b}})Liu, Li, Li, Li, Zhang, Shen, and Lee]{liu2024llavanext}
Haotian Liu, Chunyuan Li, Yuheng Li, Bo Li, Yuanhan Zhang, Sheng Shen, and Yong~Jae Lee.
\newblock Llava-next: Improved reasoning, ocr, and world knowledge, 2024{\natexlab{b}}.

\bibitem[Liu et~al.(2024{\natexlab{c}})Liu, Duan, Zhang, Li, Zhang, Zhao, Yuan, Wang, He, Liu, et~al.]{liu2024mmbench}
Yuan Liu, Haodong Duan, Yuanhan Zhang, Bo Li, Songyang Zhang, Wangbo Zhao, Yike Yuan, Jiaqi Wang, Conghui He, Ziwei Liu, et~al.
\newblock Mmbench: Is your multi-modal model an all-around player?
\newblock In \emph{European conference on computer vision}, pages 216--233. Springer, 2024{\natexlab{c}}.

\bibitem[Liu et~al.(2024{\natexlab{d}})Liu, Li, Huang, Yang, Yu, Li, Yin, lin Liu, Jin, and Bai]{ocrbench}
Yuliang Liu, Zhang Li, Mingxin Huang, Biao Yang, Wenwen Yu, Chunyuan Li, Xucheng Yin, Cheng lin Liu, Lianwen Jin, and Xiang Bai.
\newblock {OCRBench: On the Hidden Mystery of OCR in Large Multimodal Models}.
\newblock arXiv:2305.07895, 2024{\natexlab{d}}.

\bibitem[Lu et~al.(2022)Lu, Mishra, Xia, Qiu, Chang, Zhu, Tafjord, Clark, and Kalyan]{scienceqa}
Pan Lu, Swaroop Mishra, Tanglin Xia, Liang Qiu, Kai-Wei Chang, Song-Chun Zhu, Oyvind Tafjord, Peter Clark, and Ashwin Kalyan.
\newblock {Learn to Explain: Multimodal Reasoning via Thought Chains for Science Question Answering}.
\newblock In \emph{Advances in Neural Information Processing Systems}, pages 2507--2521. Curran Associates, Inc., 2022.

\bibitem[Masry et~al.(2022)Masry, Long, Tan, Joty, and Hoque]{chartqa}
Ahmed Masry, Do~Xuan Long, Jia~Qing Tan, Shafiq Joty, and Enamul Hoque.
\newblock {ChartQA: A Benchmark for Question Answering about Charts with Visual and Logical Reasoning}.
\newblock In \emph{Findings of the Association for Computational Linguistics: ACL 2022}, pages 2263--2279. Association for Computational Linguistics, 2022.

\bibitem[Mathew et~al.(2021)Mathew, Karatzas, and Jawahar]{docvqa}
Minesh Mathew, Dimosthenis Karatzas, and C.~V. Jawahar.
\newblock {DocVQA: A Dataset for VQA on Document Images}.
\newblock In \emph{Proceedings of the IEEE/CVF Winter Conference on Applications of Computer Vision (WACV)}, pages 2200--2209, 2021.

\bibitem[{OpenAI} et~al.(2023){OpenAI}, Achiam, Adler, Agarwal, Ahmad, Akkaya, Aleman, Almeida, Altenschmidt, Altman, Anadkat, et~al.]{achiam2023gpt}
{OpenAI}, Josh Achiam, Steven Adler, Sandhini Agarwal, Lama Ahmad, Ilge Akkaya, Florencia~Leoni Aleman, Diogo Almeida, Janko Altenschmidt, Sam Altman, Shyamal Anadkat, et~al.
\newblock Gpt-4 technical report.
\newblock \emph{arXiv preprint arXiv:2303.08774}, 2023.

\bibitem[Singh et~al.(2019)Singh, Natarajan, Shah, Jiang, Chen, Batra, Parikh, and Rohrbach]{textvqa}
Amanpreet Singh, Vivek Natarajan, Meet Shah, Yu Jiang, Xinlei Chen, Dhruv Batra, Devi Parikh, and Marcus Rohrbach.
\newblock {Towards VQA Models That Can Read}.
\newblock In \emph{Proceedings of the IEEE/CVF Conference on Computer Vision and Pattern Recognition (CVPR)}, pages 8317--8326, 2019.

\bibitem[Sun et~al.(2026)Sun, Wang, Mao, Wang, Mu, Zheng, Wei, and Chen]{switch_kd}
Haoyi Sun, Xiaoxiao Wang, Ning Mao, Qian Wang, Lifu Mu, Wen Zheng, Tao Wei, and Wei Chen.
\newblock {Switch-KD: Visual-Switch Knowledge Distillation for Vision-Language Models}.
\newblock In \emph{{Proceedings of the IEEE/CVF Conference on Computer Vision and Pattern Recognition (CVPR) Findings}}, pages 9316--9326, 2026.

\bibitem[Sun et~al.(2025)Sun, Zhang, Song, Guo, Chen, Zhang, Yao, Liu, and Sun]{sun2025llavauhdv3}
Shichu Sun, Yichen Zhang, Haolin Song, Zonghao Guo, Chi Chen, Yidan Zhang, Yuan Yao, Zhiyuan Liu, and Maosong Sun.
\newblock {LLaVA-UHD v3: Progressive Visual Compression for Efficient Native-Resolution Encoding in MLLMs}, 2025.

\bibitem[Touvron et~al.(2023)Touvron, Lavril, Izacard, Martinet, Lachaux, Lacroix, Rozière, Goyal, Hambro, Azhar, Rodriguez, Joulin, Grave, and Lample]{touvron2023llamaopenefficientfoundation}
Hugo Touvron, Thibaut Lavril, Gautier Izacard, Xavier Martinet, Marie-Anne Lachaux, Timothée Lacroix, Baptiste Rozière, Naman Goyal, Eric Hambro, Faisal Azhar, Aurelien Rodriguez, Armand Joulin, Edouard Grave, and Guillaume Lample.
\newblock Llama: Open and efficient foundation language models, 2023.

\bibitem[{xAI}(2024)]{xai2024grok15v}
{xAI}.
\newblock Grok-1.5 vision preview.
\newblock \url{https://x.ai/news/grok-1.5v}, 2024.

\bibitem[Yang et~al.(2025)Yang, Li, Yang, Zhang, Hui, Zheng, Yu, Gao, Huang, Lv, et~al.]{qwen3}
An Yang, Anfeng Li, Baosong Yang, Beichen Zhang, Binyuan Hui, Bo Zheng, Bowen Yu, Chang Gao, Chengen Huang, Chenxu Lv, et~al.
\newblock {Qwen3 Technical Report}.
\newblock arXiv:2505.09388, 2025.

\bibitem[Yue et~al.(2024)Yue, Ni, Zhang, Zheng, Liu, Zhang, Stevens, Jiang, Ren, Sun, et~al.]{mmmu}
Xiang Yue, Yuansheng Ni, Kai Zhang, Tianyu Zheng, Ruoqi Liu, Ge Zhang, Samuel Stevens, Dongfu Jiang, Weiming Ren, Yuxuan Sun, et~al.
\newblock {MMMU: A Massive Multi-discipline Multimodal Understanding and Reasoning Benchmark for Expert AGI}.
\newblock In \emph{Proceedings of the IEEE/CVF Conference on Computer Vision and Pattern Recognition (CVPR)}, pages 9556--9567, 2024.

\bibitem[Zhu et~al.(2024)Zhu, Chen, Shen, Li, and Elhoseiny]{zhu2023minigpt}
Deyao Zhu, Jun Chen, Xiaoqian Shen, Xiang Li, and Mohamed Elhoseiny.
\newblock {MiniGPT-4: Enhancing Vision-Language Understanding with Advanced Large Language Models}.
\newblock In \emph{The Twelfth International Conference on Learning Representations}, pages 18378--18394, 2024.

\end{thebibliography}
}

\clearpage
\setcounter{page}{1}
\renewcommand{\thepage}{S\arabic{page}}
\appendix
\numberwithin{equation}{section}
\maketitlesupplementary
\raggedbottom

\section{Geometry-aware reconstruction}
\label{sec:supp_geometry}

We detail the reconstructor. Following the main text, $H_{v,S}$ denotes the student visual states and $\hat H_{v,R}$ their dense reconstruction; the layer index is omitted for brevity.

\paragraph{Image Geometry.}
Teacher and student use different token grids and add different amounts of padding to the bottom and right of the image. For example, in our experiments, the patch size of ViT's processor is $14 \times 14$, hence the teacher's image height and width are padded to multiples of 14 pixels, whereas a student with $\rho^2=16$ requires multiples of 56 pixels. A $524\times524$ image is therefore padded to $532\times532$ for the teacher and $560\times560$ for the student, yielding $38\times38=1{,}444$ and $10\times10=100$ visual tokens, respectively. The actual token-count ratio is thus $N_{v,T}/N_{v,S}=14.44$, rather than exactly 16, because padding rounds up each dimension separately. To align their grids, we locate each teacher token center in image coordinates and express it in student-grid coordinates. This uses each model's token-cell size rather than scaling the two full padded grids to match.

\paragraph{Bilinear Upsampling.}
At each mapped location, we interpolate up to four neighboring student visual states. For each neighboring student token, we calculate the fraction of its area covered by the image itself, excluding padding. We multiply its bilinear weight by this fraction, then normalize the weights. This gives one upsampled state per teacher token and produces $U=B(H_{v,S})$. In implementation, we bilinearly sample the features multiplied by these fractions and the fractions themselves, then divide the two results. The denominator is clamped below at $10^{-6}$. Both sampling operations use zero padding and \texttt{align\_corners=False}.

\Cref{alg:supp_geometry} explains the same computation as a per-token loop; the implementation is vectorized. The argument \texttt{image} denotes the image itself, excluding padding. The helper \texttt{cell\_size} divides each model's padded image dimensions by its token-grid dimensions. Token indices start at zero, and arithmetic on two-dimensional coordinates is component-wise. The loops omit points outside the student grid, equivalent to zero-padded sampling before renormalization.

\begin{algorithm}[t]
\caption{Geometry-aware upsampling (Python-like)}
\label{alg:supp_geometry}
\begin{lstlisting}[style=geometrypython,mathescape=true,commentstyle=\color{green!50!black},breaklines=true,breakatwhitespace=true,breakindent=1em]
def coverage(cell, image):
    # cell: the area represented by one token.
    # image: the original image , no padding.
    # Return how much of the cell is original image:
    # 1 = all image; values between 0 and 1
    # mean part of the cell is padding.
    return area(cell $\cap$ image) / area(cell)

def upsample(S, T, image):
    # S: student grid with visual features.
    # T: teacher grid; only geometry is used.
    # Each grid stores its own padded sizes.
    # image: the original image, no padding.
    # Return U: same length as teacher tokens.
    for t in T:
        # Teacher token center in image pixels.
        p = (index(t) + 0.5) * cell_size(T)

        # Express this point in the student grid.
        # -0.5 makes integers token centers.
        q = p / cell_size(S) - 0.5

        # Four grid corners surrounding q;
        # skip corners outside the grid.
        value, norm = 0, 0
        for s in neighbors(S, q):
            # Multiply the horizontal and vertical interpolation weights.
            w = product(1 - abs(q - index(s)))
            # Reduce weights for partly padded cells.
            c = coverage(cell(s), image)
            w *= c
            value += w * S[s]
            norm += w

        # Normalize the interpolation weights.
        U[t] = value / max(norm, 1e-6)
    return U
\end{lstlisting}
\end{algorithm}

\paragraph{Residual Reconstruction.}
We add continuous two-dimensional sine/cosine positional embeddings $\mathrm{pos}$, computed from teacher-cell centers normalized by the teacher's padded image dimensions, and refine the upsampled states with a single bidirectional Transformer layer:
\begin{align}
 X_0 &= U+\mathrm{pos},\nonumber\\
 X_1 &= X_0+\operatorname{MHSA}(\operatorname{RMSNorm}(X_0)),\nonumber\\
 X_2 &= X_1+\operatorname{SwiGLU}(\operatorname{RMSNorm}(X_1)),\nonumber\\
 \hat H_{v,R} &= U+W_{\mathrm{out}}\operatorname{RMSNorm}(X_2).
\end{align}
The normalization modules have separate parameters. The hidden width, head count, and feed-forward width follow the student LLM. Only $W_{\mathrm{out}}$ is initialized to zero, making the initial reconstruction equal to $U$ while retaining a direct gradient path to the student. $R_{\mathrm{mid}}$ and $R_{\mathrm{last}}$ use this architecture with independent parameters. The former receives intermediate visual states, while the latter applies the same operations to student visual states after the final RMSNorm.

\section{Experiment Details}
\label{sec:supp_experiment_details}

\paragraph{PVRL Diagnosis.} We use 1,024 unique images drawn from twelve traceable LLaVA-NeXT~\cite{liu2024llavanext} sources, with 85-86 images per source. Each sample is under a maximum text length of 2,048. All models share the manifest and retain their native preprocessing.

At each block independently, the diagnostic zeros the value projections of all visual tokens, while preserving queries, keys, the attention denominator, and residual states. We compute the NLL difference between the original and intervened models on the same response tokens, averaging first over tokens within each sample and then equally over samples.

\paragraph{Training Hyperparameters.}
\begin{table}[t]
  \centering
  \caption{Training hyperparameters across different stages.}
  \label{tab:training_hyperparameters}
  \small
  \setlength{\tabcolsep}{3pt}
  \renewcommand{\arraystretch}{1.15}
  \begin{tabular}{@{}lcccc@{}}
    \toprule
    Hyperparameter & Stage-1 & Stage-1.5 & Stage-2 & KD \\
    \midrule
    Train ViT Backbone & -- & -- & -- & -- \\
    Train Projector & \checkmark & \checkmark & \checkmark & \checkmark \\
    Train Compressor & \checkmark & -- & \checkmark & \checkmark \\
    Train LLM & -- & \checkmark & \checkmark & \checkmark \\
    Train Reconstructor & -- & -- & -- & \checkmark \\
    Projector LR & $10^{-3}$ & $2\times10^{-5}$ & $2\times10^{-5}$ & $2\times10^{-5}$ \\
    Compressor LR & $10^{-4}$ & -- & $2\times10^{-5}$ & $2\times10^{-5}$ \\
    Global Batch Size & 32 & 128 & 128 & 128 \\
    Epochs & \multicolumn{4}{c}{1} \\
    Optimizer & \multicolumn{4}{c}{AdamW} \\
    Weight Decay & \multicolumn{4}{c}{0} \\
    LR Scheduler & \multicolumn{4}{c}{Cosine decay} \\
    Warmup Ratio & \multicolumn{4}{c}{0.03} \\
    DeepSpeed & \multicolumn{4}{c}{ZeRO-1} \\
    \bottomrule
  \end{tabular}
\end{table}
The training hyperparameters are summarized in Table~\ref{tab:training_hyperparameters}. During Stage-1, we freeze the ViT and LLM, optimizing the projector with a learning rate of $10^{-3}$ and the additional compressor for $16\times$ students with $10^{-4}$. Stage-1.5, Stage-2 and knowledge distillation (KD) share a learning rate of $2\times10^{-5}$. We use $\tau=2$ in computing KL divergence and $\tau=1$ for cross entropy. All stages are trained for one epoch using AdamW with cosine learning rate decay and a 3\% warmup ratio.

\paragraph{Evaluation Protocol.}
\Cref{tab:supp_benchmarks} specifies the evaluation splits, example counts, and metrics for benchmarks. All eleven benchmarks use at most 512 generated tokens. The 43 multi-image questions in MMMU are converted to labeled image grids consistently across models.

\begin{table}[t]
\centering
\caption{Evaluation splits and metrics. Counts denote evaluated questions.}
\label{tab:supp_benchmarks}
\small
\setlength{\tabcolsep}{3pt}
\begin{tabular}{@{}llrl@{}}
\toprule
Benchmark & Split & Count & Metric\\
\midrule
GQA & Test-dev balanced & 12,578 & Exact match\\
MME & Test & 2,374 & Acc. + Acc.$^{+}$\\
RWQA & Test & 765 & Accuracy\\
SQA & Test (image subset) & 2,017 & Accuracy\\
AI2D & Test & 3,088 & Accuracy\\
MMMU & Validation & 900 & Accuracy\\
MMStar & Validation & 1,500 & Accuracy\\
ChartQA & Test & 2,500 & Relaxed accuracy\\
DocVQA & Validation & 5,349 & ANLS\\
TextVQA & Validation & 5,000 & VQA accuracy\\
OCRBench & test & 1,000 & Accuracy\\
\bottomrule
\end{tabular}
\end{table}

\section{Effect of Relation-Based Distillation}
\label{sec:supp_relation_loss}

\begin{table}[t]
  \centering
  \caption{Effect of relation distillation in our adapted LLaVA-KD baseline under $\rho^2=16$.}
  \label{tab:supp_llavakd_ablation}
  \small
  \setlength{\tabcolsep}{10pt}
  \renewcommand{\arraystretch}{1.25}
  \begin{tabular}{@{}cccr@{}}
    \toprule
    $\mathcal{L}_{\mathrm{resp}}$ & $\mathcal{L}_{\mathrm{vis}}$ & $\mathcal{L}_{\mathrm{rel}}$ & $Avg_{11}$ \\
    \midrule
    \checkmark & -- & -- & 51.0 \\ 
    \checkmark & \checkmark & -- & 51.2 \\ 
    \checkmark & \checkmark & \checkmark & 50.6 \\
    \bottomrule
  \end{tabular}
\end{table}

Beyond token-wise visual semantic distillation, previous works also explore relation-based supervision to transfer structural knowledge between visual tokens. Specifically, LLaVA-KD~\cite{cai2025llava} introduces Relation Distillation (RDist) to preserve the teacher's ability to model spatial and semantic dependencies among visual tokens. It constructs visual-token self-correlation matrices $R_v^T$ and $R_v^S$ from the final-layer outputs of the teacher and student, respectively. The relation loss is defined as $\mathcal{L}_{\mathrm{rel}} = 1-\mathrm{Cos}(R_v^S,R_v^T)$, encouraging the student to reproduce the teacher's pairwise visual-token relationships. Similarly, EM-KD~\cite{em-kd} introduces Vision-Language Affinity Distillation (VLAD), which aligns the correlations between visual and textual tokens to transfer cross-modal knowledge.

To investigate the effectiveness of relation distillation under aggressive token compression, we reproduce LLaVA-KD in our setting by reconstructing dense visual tokens from the student's final-layer outputs, enabling both visual semantic and relation distillation. As shown in Table~\ref{tab:supp_llavakd_ablation}, adding $\mathcal{L}_{\mathrm{vis}}$ to response supervision improves $\mathrm{Avg}_{11}$ from 51.0 to 51.2. However, further introducing $\mathcal{L}_{\mathrm{rel}}$ reduces the score to 50.6 under $\rho^2=16$, even below the response-only baseline.

One explanation is that aggressive compression makes it difficult to preserve fine-grained relationships among visual tokens. Although reconstruction restores spatial correspondence with the teacher, it does not necessarily recover the original relational structure. Enforcing such relationships may therefore introduce additional constraints that interfere with response prediction. Based on these observations, we retain visual semantic distillation but omit relation-based losses from our proposed DVTR-KD objective.

\section{PVRL Diagnosis After Distillation}
\label{sec:pvrl_after_dist}

\begin{figure}[t]
    \centering
    \includegraphics[width=\linewidth]{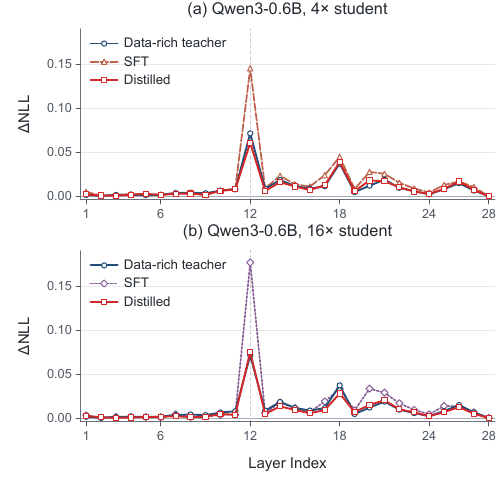}
    \caption{Layer-wise sensitivity to visual-value ablation after distillation. We compare DVTR-KD students with the data-rich teacher and the corresponding SFT students.
    }
    \label{fig:visual_readout_after_dist}
\end{figure}

We further examine the layer-wise $\Delta\mathrm{NLL}$ of students distilled with DVTR-KD, using the same images and visual-value interventions. As shown in Figure~\ref{fig:visual_readout_after_dist}, distillation substantially reduces the increase in response NLL at the PVRL for both compression ratios, bringing it closer to that of the data-rich teacher.

\end{document}